\documentclass[11pt,a4paper]{article}
\usepackage[hyperref]{conll2018}
\usepackage{times}
\usepackage{latexsym}
\usepackage{booktabs} 
\usepackage{array}
\usepackage{url}
\usepackage[subpreambles=true]{standalone}
\usepackage[normalem]{ulem}
\usepackage{amsmath}
\usepackage{amssymb}
\usepackage{tabularx}
\usepackage{soul}
\usepackage[T1]{fontenc}

\aclfinalcopy 

\newcommand\blfootnote[1]{%
  \begingroup
  \renewcommand\thefootnote{}\footnote{#1}%
  \addtocounter{footnote}{-1}%
  \endgroup
}

\title{Improving Response Selection in Multi-turn Dialogue Systems\\ 
by Incorporating
Domain Knowledge}

\author{Debanjan Chaudhuri \\
   Smart Data Analytics Group \\
 University of Bonn \& Fraunhofer IAIS \\
  Germany\\
   {\tt chaudhur@cs.uni-bonn.de} \\\And
   Agustinus Kristiadi \\
   Smart Data Analytics Group \\
   University of Bonn \\
   Germany\\
   {\tt kristiadi@uni-bonn.de}\\
   \AND 
   Jens Lehmann \\
   Smart Data Analytics Group \\
   University of Bonn \& Fraunhofer IAIS \\
   Germany \\
   {\tt jens.lehmann@cs.uni-bonn.de} \\
   \And
   Asja Fischer \\
   Department of Mathematics \\
   Ruhr University Bochum \\
   Germany\\
   {\tt asja.fischer@rub.de}\\
   }

\date{}

\begin{document}
\maketitle
\begin{abstract}
Building systems that can communicate with humans is a core problem in Artificial Intelligence. This work proposes a novel neural network architecture for response selection in an end-to-end multi-turn conversational dialogue setting. The architecture applies context level attention and incorporates additional external knowledge {provided by} descriptions of domain-specific words. It uses a bi-directional Gated Recurrent Unit (GRU) for encoding context and responses and learns to attend over the context words given the latent response representation and vice versa.In addition, it incorporates external domain specific information using {another GRU} for encoding the domain keyword descriptions. This allows better representation of domain-specific keywords in responses and hence improves the overall performance. Experimental results show that our model outperforms all other state-of-the-art methods for response selection in multi-turn conversations.
\end{abstract}

\section{Introduction}

\blfootnote{Published at the SIGNLL Conference on Computational Natural Language Learning (CoNLL) 2018.}

\begin{table}[t]
    \centering
    \renewcommand{\arraystretch}{1}
    \begin{tabularx}
    {\linewidth}
    {|>{\arraybackslash}X|}
            \hline
            \textbf{Context}\\ 
            \hline
            
            \textbf{Utterance 1:} \\
            My networking card is not working on my Ubuntu, can somebody help me? \\[3pt]
            
            \textbf{Utterance 2:} \\
            What's your kernel version? Run \textit{uname} -r or \textit{sudo} \textit{dpkg} -l \textbar \textit{grep} linux-headers \textbar \textit{grep} ii \textbar \textit{awk} '{\{\textit{print} \$3}\}' and paste the output here. \\[3pt]
            
            \textbf{Utterance 3:} \\
            It's 2.8.0-30-generic. \\[3pt]
            
            \textbf{Utterance 4:} \\
            Your card is not supported in that kernel. You need to upgrade, that's like decade old kernel! \\[3pt]
            
            \textbf{Utterance 5:} \\
            Ok how do I install the new kernel?? \\[3pt]
            
            \hline
            \textbf{Response} \\
            \hline
            Just do \textit{sudo} \textit{apt-get} upgrade, that's it.\\
            \hline
        \end{tabularx}

    \label{tab:resp}
    \caption{Illustration of a multi-turn conversation with domain specific words (UNIX commands) in italics.}
    
\end{table}


In a conversation scenario, a dialogue system can 
be applied to the task of freely generating a new response or to the task of selecting a response from a set of candidate responses based on the {previous utterances, i.e.~the}
context of the dialogue. The former is known as \emph{generative} {dialogue system} while the latter is called \emph{retrieval-based} (or response selection) dialogue system.

Both 
approaches can be realized using a modular architecture, where each module is responsible for a certain task such as natural language understanding, dialogue state-tracking, natural language generation, etc.,~or can be trained in an end-to-end manner optimized on a single objective function.

Previous 
{work}, {belonging to the latter category,} by \citet{lowe2015ubuntu} applied neural networks to multi-turn response selection in
conversations 
{by encoding} the utterances in the context {as well as the possible responses with a Long Short-term Memory (LSTM)~\cite{hochreiter1997long}}. Based on the context and response encodings, the neural network then estimates the probability for each response to be the correct one given the context. 
%
{More recently, a lot of enhanced architectures have been proposed that build on the general idea of encoding response and context first and performing some  embedding-based matching after \cite{yan2016learning, zhou2016multi,an2018improving,dong2018enhance}.}



Although such approaches result in efficient text-pair
matching capabilities, they fail to attend over logical consistencies 
for longer utterances in the context, given the response.
Moreover, in domain specific scenarios, a system's ability to incorporate additional domain knowledge 
can be very beneficial, e.g. for the example shown in Table~\ref{tab:resp}. 
In this paper, we propose a novel neural network architecture {for multi-turn response-selection} that extends the model proposed by \citet{lowe2015ubuntu}. Our major contributions are: (1) a neural network paradigm that is able to attend over important words in a context utterance 
{given} the response 
{encoding} (and vice versa), (2) 
{an approach} to incorporate additional domain knowledge into the neural network 
{by encoding the description of domain specific words with a}
GRU and {using} a bilinear operation to merge the 
{resulting domain specific representations}
with the vanilla word embeddings, and (3) 
{an empirical} evaluation on a publicly available multi-turn dialogue corpus {showing} that our system outperforms all other state-of-the-art methods for response selection in a multi-turn setting.


%


\section{Related work}

Recently, human-computer conversations have attracted increasing attention in the research community and dialogue systems have become a field of research on its own. 
The conversation models proposed in early studies    \cite{walker2001quantitative, oliver2004generating, stent2002user} were designed for catering to specific domains only, e.g.~for performing restaurant bookings, and required substantial rule-based strategy building and human efforts in the building process. With the advancements in machine learning, there have been 
{more and more}
studies on conversational agents {which are based on} data-driven approaches. 
{Data-driven dialogue 
{systems} can chiefly be realized 
{by} two types of architectures:}
 (1) {pipeline} architectures, which follow a modular pattern for modelling the dialogues, {where each component is trained/created separately to perform a specific sub-task}, and (2) end-to-end architectures, which consist of a single trainable module for modelling the conversations.

Task-oriented dialogue systems, which are designed to assist users in achieving specific goals, 
{were mainly} realized 
{by pipeline} architectures.
Recently however, there have been  
{more and more} works 
{on} end-to-end {dialogue systems}
because of {the} limitations {of} the former modular architectures, namely, the credit assignment problem and inter-component dependency, as for example described by \citet{zhao2016towards}. \citet{wen2016network} and \citet{bordes2016learning} proposed encoder-decoder-based neural 
{networks} for {modeling} task oriented dialogues. 
Moreover, \citet{zhao2016towards} proposed an end-to-end reinforcement {learning-based system}
for jointly 
{learning to perform}
dialogue state-tracking \cite{williams2013dialog} and policy learning \cite{baird1995residual}. 

Since task oriented systems primarily focus on completing a specific task, they usually do not allow free flowing, articulate conversations with the user.
{Therefore,}
there has 
been  considerable effort to develop non-goal driven dialogue systems, which are able to converse with humans on an open domain~\cite{ritter2011data}. Such systems can be modeled using either 
generative architectures, which are able to freely generate responses to user queries, or retrieval-based systems, which pick a response suitable to a {context} utterance out of a provided set of responses. Retrieval-based systems are therefore more limited in their output while having the advantage of {producing} more informative, constrained, and grammatically correct responses~\cite{ji2014information}.  
%

\subsection{Generative models}

\citet{ritter2011data} were the first to 
formulate the task of automatic response generation as phrase-based statistical machine translation, which they tackled with n-gram-based language models.
Later approaches~\cite{shang2015neural,vinyals2015neural,luong2014addressing} {applied} 
Recurrent Neural Network (RNN)-based encoder-decoder architectures.
However, dialogue generation is considerably more difficult than language translation because of the wide possibility of responses in interactions. Also, for dialogues, in order to generate a suitable response at a certain time-step, knowing only the previous utterance is often not enough and the ability to leverage the context from the sequence of previous utterances is required.
To overcome such challenges, a hierarchical RNN encoder-decoder-based system has been proposed by \citet{serban2016building} for leveraging contextual information in conversations. 

\subsection{Retrieval-based models}

Earlier works on retrieval-based systems focused on modeling short-text, single-turn dialogues.
\citet{hao2013dataset} introduced a data set for {this} task and proposed a
{response selection system which is based on information retrieval techniques like the}
vector space {model} and semantic matching.
\citet{ji2014information} {suggested to apply} a deep neural network 
{for matching} contexts and responses, {while} \citet{wu2016topic} proposed a topic aware convolutional neural tensor network for answer retrieval in short-text scenarios.

More recently, there has been a lot of focus on developing retrieval-based models for multi-turn dialogues which is more challenging as the models need to take into account long-term dependencies in the context.
\citet{lowe2015ubuntu}, introduced the Ubuntu Dialogue Corpus (UDC), 
which is the largest freely available multi-turn dialogue data set.
Moreover, the authors proposed to leverage {RNNs}, e.g.~LSTMs, to encode both the context and the response, before computing the score of the pair based on the similarity of the encodings (w.r.t.~a certain measure). This class of methods is referred to as dual encoder architectures.
Shortly after, \citet{kadlec2015improved} investigated the performance of dual encoders with
{different kind of}
encoder networks, such as convolutional neural networks (CNNs) and bi-directional LSTMs.  {\citet{yan2016learning} followed a different approach and trained a single CNN to map a context-response pair to the corresponding matching score.}

Later on, various extensions of the dual encoder architecture have been proposed.
\citet{zhou2016multi} {employed two encoders in parallel, one working on word- the other on utterance-level.} 
\citet{wu2017sequential} proposed the Sequential Matching Network (SMN), where the candidate response is matched with every utterance in the context separately, based on which a final score is computed.
{The Cross Convolution Network (CNN)} \cite{an2018improving} {extends} the dual encoder 
with a cross convolution operation. The latter is a dot product between the embeddings of the context and response followed by a max-pooling operation. Both of the outputs are concatenated and fed into a fully-connected layer for similarity matching.  Moreover, \citet{an2018improving} 
improve the representation of rare words by learning different embeddings for them from the data.
Handling rare words has also been studied by  \citet{dong2018enhance}, who proposed to handle {Out-of-Vocabulary} (OOV) words by using both pre-trained word embeddings and embeddings from task-specific data.


%
Furthermore, 
many models targeting 
response selection along with other sentence pair scoring tasks such as paraphrasing, semantic text scoring, and recognizing textual entailment have been proposed. 
\citet{baudivs2016sentence} 
{investigated a stacked RNN-CNN}
architecture and attention-based models for sentence-pair scoring.
Match-SRNN \cite{wan2016match} {employs a} spatial RNN to capture local interactions between sentence {pairs}.  Match-LSTM \cite{wang2015learning} 
improves its matching performance by {using} LSTM-based, attention-weighted sentence representations.
QA-LSTM \cite{tan2015lstm} {uses a simple attention mechanism and combines the LSTM encoder with a CNN.}

Incorporating unstructured domain knowledge into dialogue system has initially been studied {by} \citet{lowe2015incorporating} and followed by \citet{xu2016incorporating}, who incorporated a loosely-structured 
knowledge base into a neural network using a special gating mechanism. They created the knowledge base from 
domain-specific data, however their model is not able to leverage any external 
domain knowledge.

\section{Background}

In {this} section, we will explain the task at hand and give a brief introduction to the neural network architectures our proposed model is based on.

\subsection{Problem definition}

Let the data set $\mathcal{D} = \{(c_i, r_i, y_i)\}_{i=1}^M$ be a 
{set} of $M$ triples {consisting} of context $c_i$, response $r_i$, and ground truth label $y_i$. Each context is a sequence of utterances, that is $c_i = \{ u_{il} \}_{l=1}^L$, where $L$ is the maximum context length. We define a{n} utterance as a sequence of words $\{ w_t \}_{t=1}^T$. Thus,  $c_i$ can also be viewed as a sequence of words by concatenating all utterances in $c_i$. Each response $r_i$ is an utterance and $y_i \in \{0, 1\} $ is the corresponding label of the given triple which takes a value of 1 if $r_i$ is the correct response for $c_i$ and 0 otherwise. 
The goal of retrieval-based dialogue systems is then 
to learn a predictive distribution $p(y \vert c, r, \boldsymbol{\theta})$ parameterized by $\boldsymbol{\theta}$. That is, given a context $c$ and response $r$, we would like to infer 
the probability of $r$ being a response to
context $c$. 

\subsection{RNNs, BiRNNs and GRUs}
\label{sec:RNN}


Recurrent neural networks 
are one of the most popular classes of models for processing sequences of words  $W = \{w_t\}_{t=1}^T$ with arbitrary length $T \in \mathbb{N}$, e.g.~utterances or sentences.
%
%
Each word $w_t$ is first mapped onto its vector representation $\mathbf{w}_t$ (also referred to as word embedding), which serves as input to the RNN at 
{time step} $t$.
The central {element} of RNNs is the recurrence relation of its hidden units, described by
\begin{equation}
    \overrightarrow{\mathbf{h}}_t = f(\overrightarrow{\mathbf{h}}_{t-1}, \mathbf{w}_t \vert \boldsymbol{\phi}) \enspace ,
\end{equation}
\noindent 
where $\boldsymbol{\phi}$ are the parameters of the RNN and $f$ is some nonlinear function.
Accordingly, the 
state $\overrightarrow{\mathbf{h}}_{t}$ of the hidden units {at time step $t$}
depends on the state $\overrightarrow{\mathbf{h}}_{t-1}$ in the previous time step  
and the $t$-th word in the sequence. 
This way, the hidden state $\overrightarrow{\mathbf{h}}_{T}$ obtained after $T$ updates contains information about the whole sequence $W$, and can thus be regarded as an embedding of the sequence.

The RNN architecture can also be altered to take into account dependencies coming from both the past and the future by adding an additional sub-RNN that moves backward in time, giving rise to the name bi-directional RNN (biRNN).
%
{To achieve this}, the network architecture is extended by an additional set of hidden units. 
The states {$\overleftarrow{\mathbf{h}_t}$} 
{of those hidden units} are updated based on the current input word and the hidden state from the next time step. That is for $t=1,\dots,{T-1}$:
\begin{equation}
   \overleftarrow{\mathbf{h}}_{T-t} = f( \overleftarrow{\mathbf{h}}_{T-t+1}, \mathbf{w}_{T-t} \vert {\boldsymbol{\phi}}) \enspace.
\end{equation}
\noindent Here, the words are processed in 
reverse order, i.e.~$w_T,\dots,w_1$, such that {$\overleftarrow{\mathbf{h}}_{T}$ {(analogous to $\overrightarrow{\mathbf{h}}_{T}$ in the forward directed RNN)} contains information about the whole sequence.
At the $t$-th time step, the model's hidden representation of the sequence is then usually {obtained} by the concatenation of the hidden 
{states} from the forward and the backward RNN, i.e.~by 
{$\mathbf{h}_t=[\overrightarrow{\mathbf{h}}_{t}, \overleftarrow{\mathbf{h}}_{t}]$}
and the embedding of the whole sequence $W$ is given by 
{$\mathbf{h}^{W}=
[\overrightarrow{\mathbf{h}}_{T}, \overleftarrow{\mathbf{h}}_{T}]$.}


Modeling very long sequence{s} with RNNs is hard: \citet{bengio1994learning} showed that RNNs suffer from vanishing and exploding gradients, which makes training over long-term dependency difficult. Such problems can be addressed by augmenting the RNN with additional gating mechanisms, as {it} is done in 
{LSTMs} and the Gated Recurrent Unit (GRU)~\cite{cho2014gru}. These mechanisms allow the RNN to learn how much to update the hidden state flexibly in each step and help the RNN to deal with the vanishing gradient problem in long sequences better than vanilla RNNs. The gating mechanism of GRUs is motivated by that of LSTMs, but is much simpler to compute and implement. 
It contains two gates, namely the reset and update {gate}, 
{whose states} at time $t$ {are} denoted by {$\mathbf{z_t}$ and $\mathbf{r_t}$}, respectively. Formally, {a} GRU is defined {by the following update equations}
\noindent
\begin{align*}
    \mathbf{z}_t &= \sigma({\mathbf{W}_z}{\mathbf{x}_{t}} + \mathbf{U}_z \mathbf{h}_{t-1})\enspace,\\
    \mathbf{r}_t &= \sigma(\mathbf{W}_r \mathbf{x}_{t} + \mathbf{U}_r \mathbf{h}_{t-1})\enspace,\\
    {\Tilde{\mathbf{h}}_t} &= \text{tanh}(\mathbf{W}_h \mathbf{x}_t + \mathbf{U}_h \mathbf{r}_i \odot \mathbf{h}_{t-1}) \enspace,\\ 
    \mathbf{h}_t &= \mathbf{z}_t \odot \Tilde{\mathbf{h}}_t + (1-\mathbf{z}_t) \odot \mathbf{h}_{t-1}\enspace,
\end{align*}
\noindent
{where}
$\mathbf{x}_t$ is the input  {(corresponding to $\mathbf{w}_t$ in our setting)} and {the set of weight matrices $\mathbf{\phi}=\{\mathbf{W}_z$, $\mathbf{U}_z$,$\mathbf{W}_r$, $\mathbf{U}_r$, $\mathbf{W}_h$, $\mathbf{U}_h\}$ constitute the learnable model parameters.}
%




\subsection{Dual Encoder}
\label{sec:RNNDualEncoder}

Recurrent neural networks and their variants have been used in many applications in the field of natural language processing{, including retrieval-based dialogue systems}. 
In this area {the} 
dual encoder (DE)~\cite{lowe2015ubuntu} became a popular model. It uses a single RNN encoder to transform both context and response into low dimensional vectors and computes their similarity.
More formally,  
let $\mathbf{h^c}$ 
and $\mathbf{h^r}$
be the encoded context and response, respectively. The probability of $r$ being the correct response {for} $c$ 
is then computed by the DE as
\begin{align}
\label{eq:DE}
   p(y \vert c, r, \boldsymbol{\theta}) 
    =\sigma((\mathbf{h^c})^\text{T} \, \mathbf{M} \, \mathbf{h^r} + b) \enspace ,
\end{align}
\noindent 
where $\boldsymbol{\theta} = \{ \boldsymbol{\phi}, \mathbf{M}, b \}$
{(recall, that $\boldsymbol{\phi}$ is the set of parameters of the encoder RNN that outputs $\mathbf{h^c}$ and $\mathbf{h^r}$ )}
is the set of parameters {of the full model} and $\sigma$ is the sigmoid function. 
Note, that the same RNN is used to encode both context and response.

In summary, 
this approach can be described as first creating latent 
representations of context and response in the same vector space and then using the {similarity} between these latent embeddings 
(as
induced by matrix $\mathbf{M}$ and bias $b$)
for estimating
the probability 
of the 
{the response being the correct one for}
the given context. 



\section{Model description}

Our model extends the DE described  in  Section~\ref{sec:RNNDualEncoder} by two attention mechanisms which make the context encoding response-aware and vice {versa}.
Furthermore, we augment the model with a mechanism for incorporating external knowledge to improve the handling of rare words. Both extensions are described in detail in the following subsections.




\subsection{{Attention augmented encoding}}

As described above, in the 
{DE} context and response are encoded independently from each other based on the same RNN. Instead of simply taking the final hidden state $\mathbf{h}^c$ (and $\mathbf{h}^r$) of the RNN as context (and response) encoding,
we propose to use a response-aware attention mechanism to calculate the context embedding and vice versa.


Subsequently, we 
{will} describe this mechanism formally. Recall that a context $c$ can be seen as sequence of words $\{ w^c_{t} \}_{t=1}^T$  where all utterances are concatenated and $T$ is the total number of words in the context. 
{Given} this 
{sequence}, the RNN 
{(in our experiments a bi-directional GRU)}
produces a sequence of hidden states $\mathbf{h}^c_1,\dots, \mathbf{h}^c_T$ and an encoding of the whole context sequence $\mathbf{h}^c$ as described in Section~\ref{sec:RNN}. 
Analogously, we get $\mathbf{h}^r_1,\dots, \mathbf{h}^r_{T'}$ and $\mathbf{h}^r$ for a response consisting of a sequence of words $\{w^r_t\}^{T'}_{t=1}$, where $T'$ is the total number of words in the response.



For calculating the response-aware context encoding, we first estimate attention weights $\alpha_{t}^c$  for the hidden state $\mathbf{h}^c_t$ in each time step, depending on the response encoding $\mathbf{h}^r$:
\begin{align}
    \alpha_{t}^c \propto \exp(({\mathbf{h}_{t}^c})^\text{T}{\mathbf{W}_{c}}{\mathbf{h}^r})\enspace,
\end{align}
where $\mathbf{W}_{c}$ is a learnable parameter matrix.
The response-aware context embedding {then} is 
given by
\begin{align}
    \hat{\mathbf{h}}^{c} = \sum_{t=1}^T \alpha^c_t \, \mathbf{h}_t^c \enspace.
\end{align}
Intuitively this means, that  depending on the response we focus on different parts of the context sequence, for judging on how well the response matches the context. 
{This may resemble human focus.} 

Similarly, 
{we calculate the context-aware response encoding by 
\begin{align}
        \hat{\mathbf{h}}^{r} =\sum_{t=1}^T \alpha^r_t \, \mathbf{h}_t^r \enspace,
\end{align}
with attention weights
\begin{align}
    \alpha_{t}^r \propto \exp({(\mathbf{h}_{t}^r)}^\text{T}{\mathbf{W}_{r}}{\mathbf{h}^c}) \enspace.
\end{align}
}
%
The two attention-weighted encodings (for  response and context, respectively)} then replace the vanilla encodings in equation \eqref{eq:DE}, that is
\begin{align}
   p(y \vert c, r, \boldsymbol{\theta}) 
    =\sigma((\hat{\mathbf{h}}^c)^\text{T} \, \mathbf{M} \, \hat{\mathbf{h}}^r + b) \enspace .
\end{align}


\subsection{Incorporating domain keyword descriptions}

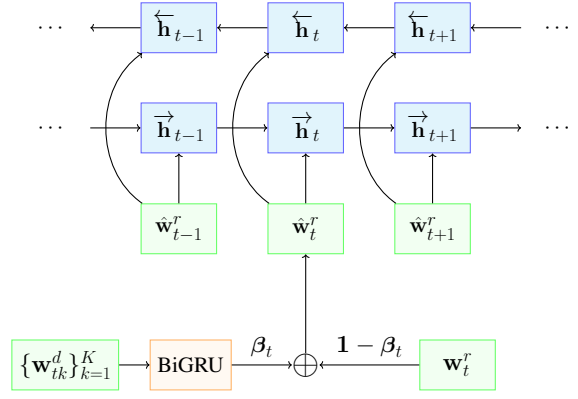
\begin{figure}[t]
      \begin{tikzpicture}[scale=3, node distance=2cm, auto, align=center, every node/.style = {draw=blue!60, fill=cyan!10, minimum height=2cm, minimum width=3cm}]

    \Huge
    
    \node[draw=none, fill=none] (h_t-2) {$\cdots$};
    \node(h_t-1) [draw, very thick, right=of h_t-2] {$\overrightarrow{\mathbf{h}}_{t-1}$};
    \node(h_t)[draw, very thick, right=of h_t-1] {$\overrightarrow{\mathbf{h}}_{t}$};
    \node(h_t+1) [draw, very thick, right=of h_t] {$\overrightarrow{\mathbf{h}}_{t+1}$};
    \node(h_t+2) [draw=none, fill=none, right=of h_t+1] {$\cdots$};
    
    \draw[very thick, ->] (h_t-2) -- (h_t-1);    
    \draw[very thick, ->] (h_t-1) -- (h_t);
    \draw[very thick, ->] (h_t) -- (h_t+1);
    \draw[very thick, ->] (h_t+1) -- (h_t+2);
    
    \node[draw=none, fill=none, above=of h_t+2] (g_t-2) {$\cdots$};
    \node(g_t-1) [draw, very thick, left=of g_t-2] {$\overleftarrow{\mathbf{h}}_{t+1}$};
    \node(g_t)[draw, very thick, left=of g_t-1] {$\overleftarrow{\mathbf{h}}_{t}$};
    \node(g_t+1) [draw, very thick, left=of g_t] {$\overleftarrow{\mathbf{h}}_{t-1}$};
    \node(g_t+2) [draw=none, fill=none, left=of g_t+1] {$\cdots$};
    
    \draw[very thick, ->] (g_t-2) -- (g_t-1);    
    \draw[very thick, ->] (g_t-1) -- (g_t);
    \draw[very thick, ->] (g_t) -- (g_t+1);
    \draw[very thick, ->] (g_t+1) -- (g_t+2);

    \node(e_t-1) [draw=green!60, fill=green!5, very thick, below=of h_t-1] {$\hat{\mathbf{w}}^r_{t-1}$};
    \node(e_t) [draw=green!60, fill=green!5, very thick, below=of h_t] {$\hat{\mathbf{w}}^r_{t}$};
    \node(e_t+1) [draw=green!60, fill=green!5, very thick, below=of h_t+1] {$\hat{\mathbf{w}}^r_{t+1}$};
    
    \draw[very thick, ->] (e_t-1) -- (h_t-1);
    \draw[very thick, ->] (e_t) -- (h_t);
    \draw[very thick, ->] (e_t+1) -- (h_t+1);
    
    \draw[very thick, bend left=55, ->] (e_t-1) to (g_t+1);
    \draw[very thick, bend left=55, ->] (e_t) to (g_t);
    \draw[very thick, bend left=55, ->] (e_t+1) to (g_t-1);

    \node(plus) [circle, cross=none, draw=none, fill=none, minimum size=0cm, below=of e_t, yshift=-5ex] {\fontsize{40pt}{60pt}\selectfont $\oplus$};
    \node(e_t_ori) [draw=green!60, fill=green!5, very thick, right=of plus, xshift=5ex] {$\mathbf{w}^r_t$};
    \node(cnn) [draw=orange!60, fill=orange!5, rectangle, very thick, left=of plus, xshift=-1ex] {BiGRU};
    \node(desc_embeds) [draw=green!60, fill=green!5, very thick, left=of cnn, xshift=2ex] {$\{ \mathbf{w}^d_{tk} \}_{k=1}^K$};
    
    \draw[very thick, ->] (e_t_ori) -- (plus) node[draw=none, fill=none, minimum size=0, midway, above] {$\mathbf{1}-\boldsymbol{\beta}_t$};
    \draw[very thick, ->] (plus) -- (e_t);
    \draw[very thick, ->] (cnn) -- (plus) node[draw=none, fill=none, minimum size=0, midway, above] {$\boldsymbol{\beta}_t$};
    \draw[very thick, ->] (desc_embeds) -- (cnn);

  \end{tikzpicture}
   
    \label{fig:incorporating_desc}
    \caption{Our proposed way to incorporate domain knowledge into the model. $\boldsymbol{\beta}_t$ and $\mathbf{1} - \boldsymbol{\beta}_t$ represent the (multiplicative) weights for the description embedding and the word embedding respectively. The resulting combination, $\hat{\mathbf{w}}^r_t$ acts as an input of the encoder.}
\end{figure}


\citet{bahdanau2018learning} proposed a method for learning embeddings for {OOV} words 
{based on} external dictionary definitions. They learn {these} description embeddings of words using an LSTM {for encoding the corresponding definition}. If a particular word included in the dictionary also appears in the corpus' vocabulary 
{(for which vanilla word embeddings are given)}, they add the word embedding and the description embedding 
{together}. Otherwise, 
{in the case of OOV words}, they use solely the description embedding in place of the missing word embedding. 
Inspired by this approach, we use a similar technique to incorporate domain keyword descriptions into 
word embeddings.


If a word $w^r_t$ in the response utterance is in the set of domain keywords $\mathcal{K}$, we firstly extract its description. The description of $w^r_t$ is a sequence of words $\{ w^d_{tk} \}_{k=1}^K$, which {is} projected {onto} sequence of embeddings $\{ \mathbf{w}^d_{tk} \}_{k=1}^K$.
This sequence is 
encoded using another bi-directional GRU 
to obtain a vector representation $\mathbf{h}^d_t$ 
{of the same dimension as the vanilla word embeddings.}
If $w^r_t$ is not in $\mathcal{K}$, 
{we simply}
set $\mathbf{h}^d_t$ to zero.
We call $\mathbf{h}^d_t$ {the} \emph{description embedding}.

Some 
domain specific words might also happen to be common words. For instance, 
in the case of the UDC's vocabulary, there 
exist tokens such as \textit{shutdown}~\footnote{UNIX command for system shutdown.} or \textit{who}~\footnote{UNIX command to get {a list of} currently logged-in users.},
which are ambiguous, {i.e.,} although they are valid UNIX commands, they are also common words in natural language.
The description embeddings of domain specific words can be simply added to the vanilla word embeddings as suggested by \citet{bahdanau2018learning}.{
However, it might be advantageous if the model can determine itself whether to treat the current word as a domain specific word, {a} common word, or 
{something} in between, depending on the context.
For instance, if the context is mainly talking about system users, then \textit{who} is most likely a UNIX keyword.}
{{Therefore}, we propose a more flexible way to combine 
the description embedding $\mathbf{h}^d_t$ and the word embedding $\mathbf{w}^r_t$}, that is,
{we define the final word embedding to be a {convex} combination of both, 
and let
the combination coefficients be given by a function of $\mathbf{h}^d_t$ and the context embedding $\hat{\mathbf{h}}^c$.} 
{Intuitively, this allows the model to flexibly focus on the description or the vanilla embedding, in dependence on the context and the description.}
{Formally, the 
{combination coefficients}
$\boldsymbol{\beta}_t$ of t-th word in the response is given by}
\begin{align}
    \boldsymbol{\beta}_t \propto \exp(\mathbf{U}^\text{T} \hat{\mathbf{h}}^c + \mathbf{V}^\text{T} \mathbf{w}^r_t) \enspace ,
\end{align}
where $\mathbf{U}$ and $\mathbf{V}$ are learnable parameter matrices. Note that $\boldsymbol{\beta}_t$ is a vector 
{of the same dimension} as the embeddings.
The final embedding {of $w^r_t$ ({which serves as input to the response encoder}) is then the weighted sum} 
\begin{align}
    \hat{\mathbf{w}}^r_t = \boldsymbol{\beta}_t \odot \mathbf{h}^d_t + (\mathbf{1} - \boldsymbol{\beta}_t) \odot \mathbf{w}^r_t \enspace ,
    \label{eq:keweight}
\end{align}
{where $\odot$ denotes the element wise multiplication.}





\section{Experiment}

\subsection{Ubuntu multi-turn {d}ialogue {c}orpus}

Extending the 
{work} of \citet{uthus2013extending}, \citet{lowe2015ubuntu} introduced a version of the Ubuntu chat log conversations which is the largest publicly available
multi-turn, dyadic, and domain-specific {dialogue} data set. 
The chats are extracted from Ubuntu related topic specific chat rooms in the Freenode Internet Relay Chat (IRC) network. 
{Usually, experienced users address a problem of someone by suggesting a potential solution and a \textit{name mention} of the addressed user.} 
A conversation between a pair of users {often} stops when the problem has been solved. However, they might continue having a discussion which is not related to the topic.

A preprocessed {version of the} above corpus and {the needed} vocabulary are provided by \citet{wu2017sequential}. The preprocessing {consisted} of replacing numbers, URLs, and system paths with special placeholders as suggested by \citet{xu2016incorporating}.  No additional preprocessing 
is performed by us.
{The data set} consists of 1 million training triples, 500k validation triples, and 500k test triples.
{One half of the 1 million training triples are positive 
(triples with $y = 1$, i.e.~the provided response fits the context) the other half negative (triples with $y = 0$)}. 
{In contrast}, in the validation and test set, for every context $c_i$, {there exists} one positive triple 
{providing the}
ground-truth response {to} $c_i$ and nine negative triples
with  unbefitting responses.
Thus, in these sets, the ratio between positive and negative triples {per context} is 1:9 
{which makes} evaluating the model with information retrieval metrics such as Recall@k  {possible} {(see Section \ref{sec:results})}.


\begin{table*}[ht]
    \centering
    
    \begin{tabular}{ l c c c c  } 
        \toprule
        
        \textbf{Model} & $\mathbf{R_2@1}$ & $\mathbf{R_{10}@1}$ & $\mathbf{R_{10}@3}$ & $\mathbf{R_{10}@5}$\\
        
        \toprule
        
        {DE}-RNN~\cite{kadlec2015improved} & 0.768 & 0.403 & 0.547 &0.819 \\ 
         {DE}-CNN~\cite{kadlec2015improved} & 0.848 & 0.549 & 0.684 & 0.896 \\ 
         {DE}-LSTM~\cite{kadlec2015improved} & 0.901 & 0.638  & 0.784 & 0.949 \\
         {DE}-BiLSTM~\cite{kadlec2015improved} & 0.895 & 0.630 & 0.780 & 0.944 \\
        \cmidrule{1-5}
        
        MultiView~\cite{zhou2016multi} & 0.908 & 0.662 & 0.801  & 0.951\\
        DL2R~\cite{yan2016learning} & 0.899 & 0.626 &0.783 &0.944\\
        r-LSTM~\cite{xu2016incorporating} & 0.889 & 0.649& 0.857 & 0.932\\
        \cmidrule{1-5}
        
        MV-LSTM~\cite{wan2016match} & 0.906 & 0.653 & 0.804 & 0.946\\
        Match-LSTM~\cite{wang2015learning} & 0.904 &  0.653  &  0.799  &0.944\\
        QA-LSTM~\cite{tan2015lstm} & 0.903 & 0.633 & 0.789 & 0.943\\
        
        \cmidrule{1-5}
        
        SMN$_{\text{dyn}}$~\cite{wu2017sequential} & 0.926& 0.726& 0.847& 0.961 \\
        CCN~\cite{an2018improving} &  - & 0.727 & 0.858 & 0.971\\
        \hline
        ESIM~\cite{dong2018enhance} &  - & 0.734 & 0.854 & 0.967 \\
        
        \cmidrule{1-5}
        
        AK-DE-biGRU (Ours) & \textbf{0.933} & {\textbf{0.747}} & {\textbf{0.868}} & \textbf{0.972}\\
        
        \bottomrule
    \end{tabular}
    
    \caption{Evaluation results of our models compared to various baselines on Ubuntu Dialogue Corpus.} 
    \label{tab:ubuntu}
\end{table*}

\subsection{Model hyperparameters}
We 
{chose a} word {embedding} dimension of  200 as {done by} \citet{wu2017sequential}. We use fastText \cite{bojanowski2016enriching} to pre-train the word embeddings using the training set instead of using off-the-shelf word embeddings, following~\citet{wu2017sequential}. We set the hidden dimension of our GRU to be 300, as in {the work of} \citet{lowe2015ubuntu}. {We restricted the} sequence length of a context by a maximum of 320 
words, and that of the response {by} 160.
Because of the resulting size of the model and limited GPU memory, we had to use a smaller batch size of 32. We optimize the binary cross entropy loss of our model with respect to the training data using Adam~\cite{kingma2014adam} with  
{an initial} learning rate of 0.0001.
We train our model for a maximum of 20 epochs as according to our experience, this is more than enough to achieve convergence. The training is stopped when the validation recall does not increase after three subsequent epochs. The test set is evaluated on the model with the best validation recall.

For the implementation, we use PyTorch \cite{paszke2017automatic}. We train the model end-to-end with a single 12GB GPU.
The implementation\footnote{https://github.com/SmartDataAnalytics/AK-DE-biGRU.} of our models along with the additional domain knowledge base\footnote{Command descriptions scraped from Ubuntu man pages.} are publicly available.

\section{Results}
\label{sec:results}


Following \citet{lowe2015ubuntu} and \citet{kadlec2015improved}, we use {the} Recall@k evaluation metric, 
{where $R_n@k$ corresponds to the fraction of of examples for which the correct response is under the $k$ best out of a set of $n$ candidate responses, which were ranked according to there their probabilities under the model.}

In our evaluation specifically, we use $R_{2}@1$, $R_{10}@1$, $R_{10}@3$, and $R_{10}@5$.


\subsection{Comparison against baselines}

{We compare our model, which we refer to as \emph{Attention and external Knowledge augmented DE with bi-directional GRU} (\textbf{AK-DE-biGRU}), against models previously tested on the same data set:}
%
%
%
   the basic DE models
    {analyzed} by \citet{lowe2015ubuntu} and \citet{kadlec2015improved} 
    {using different encoders, such as 
    convolutional neural network  (\textbf{DE-CNN}), LSTM (\textbf{DE-LSTM}) and bi-directional LSTM (\textbf{DE-BiLSTM)}};
 %
    %
   the \textbf{Multi-View}, \textbf{DL2R} and \textbf{r-LSTM} models proposed by \citet{zhou2016multi}, \citet{yan2016learning} and \citet{xu2016incorporating}, respectively;
    %
    {architectures for advanced context and response matching, namely}
    \textbf{MV-LSTM} \cite{wan2016match}, \textbf{Match-LSTM} \cite{wang2015learning}, and \textbf{QA-LSTM} \cite{tan2015lstm};
    %
    %
    architectures processing the context utterances individually, namely 
    \textbf{SMN$_{\text{dyn}}$}  \cite{wu2017sequential} and
     \textbf{CCN};
    and we also use recently proposed \textbf{ESIM}  \cite{dong2018enhance} as a baseline.

The results 
are reported in Table \ref{tab:ubuntu}.
Our model outperforms all other models used as baselines. 
The largest improvement of our model compared to the best of the baselines (i.e.~ESIM {in general and SMN$_\text{dyn}$ for $R_2@1$ metric}) are {with respect to the} $R_{10}@1$ and $R_{10}@3$ metric, where we observed {absolute} improvements of 0.013 and 0.014 {corresponding to 1.8\% and 1.6\% relative improvement} 
, respectively. For $R_{2}@1$ and $R_{10}@5$ 
we observed more modest improvements of 0.007 (0.8\%) and 0.005 (0.5\%), respectively.
Our results are {significantly better} with $p < 10^{-6}$  for a one-sample one-tailed t-test compared to the best baseline (ESIM), on $R_{10}@1$, $R_{10}@3$, $R_{10}@5$ metrics, using {the outcome of 15 independent experiments}. {The variance between different trials is smaller than 0.001 for all evaluation metrics.}


\subsection{Ablation study}

\begin{table}[t]
    \vspace{-0.3cm}
    \centering
    \setlength\tabcolsep{0.1cm}
    \begin{tabular}{lccc}
    \toprule
    \textbf{Model} & $\mathbf{R_{10}@1}$ & $\mathbf{R_{10}@3}$ & $\mathbf{R_{10}@5}$\\
    \toprule
        DE-GRU & 0.685 & 0.831 & 0.960 \\
        DE-biGRU & 0.678 & 0.813 & 0.956\\
        A-DE-GRU & 0.712 & 0.845 & 0.964 \\
        A-DE-biGRU & 0.739 & 0.864 & 0.968 \\
        AK$_{+}$-DE-biGRU & 0.743 & 0.867 & 0.969 \\
        AK-DE-biGRU$_{w2v}$ & 0.745 & 0.866 & 0.970\\
        AK-DE-biGRU & \textbf{0.747} & \textbf{0.868} & \textbf{0.972}\\
    \bottomrule
    \end{tabular}
    \caption{Ablation study with different settings.}
    \label{tab:ablation}
\end{table}

\noindent
{Our model differs in various ways from the vanilla DE: it uses a GRU instead of an LSTM for the encoding, introduces an attention mechanism for the encoding of the context and {another} for the encoding of the response, and incorporates additional knowledge in the response encoding process.}

To analyze {the effect of these components on the over all performance, we analyzed different model variants:
a DE using a GRU or a bi-directional GRU as encoder (\textbf{DE-GRU} and \textbf{DE-biGRU}, respectively) and both of these models with attention augmented encoding for embedding {both} context and response (\textbf{A-DE-GRU} and \textbf{A-DE-biGRU}, respectively).
We also tested the effects of using 
a simple addition instead of 
the weighted summation given in equation~\eqref{eq:keweight} for merging the word embedding with the desciption embedding (\textbf{AK$_{+}$-DE-biGRU}). Finally, we investigated a version of our model (\textbf{AK-DE-biGRU$_{w2v}$}) where we used pre-trained word2vec embeddings, as done by \citet{wu2017sequential}, instead of learning our own word embeddings from the data set.
}



The results of the study are presented in Table~\ref{tab:ablation}. With the basic models, i.e.~DE-GRU and DE-biGRU, as baselines, we observed around 4\% and 9\% improvement on $R_{10}@1$ when incorporating the attention mechanism (A-DE-GRU and A-DE-biGRU, respectively).

When domain knowledge {is} incorporated by simple addition {(as in the work of \citet{bahdanau2018learning})}, i.e.~in AK$_+$-DE-biGRU, we noticed 0.5\% further improvement. 
{Note however, that the results are not as good as when using the proposed weighted addition.}
Finally, using our method of incorporating domain knowledge {in combination} with embeddings trained from scratch with fastText~\cite{bojanowski2016enriching}, {the} {performance gets 0.3\% better than when using}
{pretrained}  word2vec 
{embeddings}.
In total, compared to {the} DE-biGRU baseline, our model (AK-DE-biGRU) achieves 10\% of improvement {in terms of the} $R_{10}@1$ metric. Thus, 
{the results clearly suggest that}
both the attention mechanism and the incorporation of domain knowledge, are effective {approaches} for improving the dual encoder architecture.
Curiously, we noticed that for the baseline models, using {a} GRU as the encoder is better than using {a} biGRU. 
{This}
finding is in line with the results from~\citet{kadlec2015improved} {reported} in Table~\ref{tab:ubuntu}. However, the table is turned when  
{augmenting the models with an attention mechanism where}
{the} biGRU-based model outperforms {the one with the} GRU. This observation motivates us to consider {a} biGRU instead of {a} GRU in our final model.


\subsection{Visualizing response attentions}

\begin{table}[t!]
    \centering
    \renewcommand{\arraystretch}{1}
    \begin{tabularx}{\linewidth}{>{\raggedright\arraybackslash}X}
    \toprule
        \textbf{Example Response Utterances} \\
    \toprule
        \colorbox{blue!17}{gui}\colorbox{blue!2}{for}\colorbox{blue!18}{shutdown}\colorbox{blue!1}{try}\colorbox{blue!9}{typing}\colorbox{blue!8}{{$sudo$}} \colorbox{blue!30}{{$shutdown$}}\colorbox{blue!8}{-h}\colorbox{blue!0}{now} \\
     \cmidrule{1-1}
        \colorbox{blue!8}{$sudo$}\colorbox{blue!6}{$apt$-$get$}\colorbox{blue!7}{{install}}\colorbox{blue!30}{qt4-designer}
        \colorbox{blue!5}{there}\colorbox{blue!4}{could}\colorbox{blue!2}{be}\colorbox{blue!5}{some}\colorbox{blue!14}{qt}\colorbox{blue!1}{dev}\colorbox{blue!1}{packages}\colorbox{blue!1}{too}
        \colorbox{blue!0}{but}\colorbox{blue!0}{i}\colorbox{blue!1}{think}\colorbox{blue!0}{the}\colorbox{blue!2}{above}\colorbox{blue!0}{will}\colorbox{blue!1}{install}\colorbox{blue!2}{them}\colorbox{blue!1}{as}
        \colorbox{blue!3}{dependencies} \\
    \cmidrule{1-1}
        \colorbox{blue!2}{certainly}\colorbox{blue!1}{won}\colorbox{blue!0}{n't}\colorbox{blue!3}{make}\colorbox{blue!2}{a}\colorbox{blue!2}{difference}\colorbox{blue!2}{i}\colorbox{blue!0}{m} \colorbox{blue!1}{sure}\colorbox{blue!0}{but}\colorbox{blue!1}{maybe}\colorbox{blue!0}{try}\colorbox{blue!5}{{$sudo$}}\colorbox{blue!15}{{$shutdown$}}\colorbox{blue!1}{-r}
        \colorbox{blue!0}{now}\colorbox{blue!32}{shutdown}\colorbox{blue!1}{works}\colorbox{blue!0}{just}\colorbox{blue!1}{fine}\colorbox{blue!5}{graphical}
        \colorbox{blue!1}{and}\colorbox{blue!5}{command}\colorbox{blue!6}{line} \\
    \cmidrule{1-1}
        \colorbox{blue!16}{pci}\colorbox{blue!0}{can}\colorbox{blue!0}{you}\colorbox{blue!1}{put}\colorbox{blue!1}{the}\colorbox{blue!8}{output}\colorbox{blue!2}{of}\colorbox{blue!61}{{$lspci$}} \colorbox{blue!0}{on}\colorbox{blue!2}{\_\_url\_\_}\colorbox{blue!1}{and}\colorbox{blue!1}{give}\colorbox{blue!0}{me}\colorbox{blue!0}{the}\colorbox{blue!4}{link}\colorbox{blue!1}{please} \\
    \cmidrule{1-1}
         \colorbox{blue!0}{i}\colorbox{blue!0}{do}\colorbox{blue!0}{n't}\colorbox{blue!0}{see}\colorbox{blue!0}{a}\colorbox{blue!1}{line}\colorbox{blue!0}{in}\colorbox{blue!17}{xorg}\colorbox{blue!1}{conf}\colorbox{blue!0}{for} \colorbox{blue!24}{hsync}\colorbox{blue!0}{and}\colorbox{blue!19}{vsync}\colorbox{blue!0}{do}\colorbox{blue!0}{you}\colorbox{blue!1}{get}\colorbox{blue!0}{the}\colorbox{blue!0}{same} \colorbox{blue!0}{you}\colorbox{blue!0}{d}\colorbox{blue!2}{create}\colorbox{blue!1}{it}\colorbox{blue!0}{i}\colorbox{blue!0}{m}\colorbox{blue!0}{looking}\colorbox{blue!0}{at}\colorbox{blue!10}{gentoo} \colorbox{blue!1}{and}\colorbox{blue!3}{ubuntu} \colorbox{blue!5}{forums}\colorbox{blue!0}{a}\colorbox{blue!3}{sec} \\
    \cmidrule{1-1}
        \colorbox{blue!1}{can}\colorbox{blue!1}{be}\colorbox{blue!1}{many}\colorbox{blue!2}{reasons}\colorbox{blue!1}{of}\colorbox{blue!50}{${traceroute}$} \colorbox{blue!23}{\_\_url\_\_}\colorbox{blue!0}{you}\colorbox{blue!0}{will}\colorbox{blue!2}{not}\colorbox{blue!2}{get}\colorbox{blue!2}{a}\colorbox{blue!4}{complete}\colorbox{blue!5}{result} \\
    \bottomrule
    \end{tabularx}
    
    \caption{{Visualization of attention weight in utterance samples, darker shade means higher attention weight.}}
    \label{tab:attention_visualization}
    \vspace{-0.3cm}
\end{table}



To further investigate the results given by our model, we qualitatively inspected several samples of response
utterances and their attention weights, as shown in 
Table~\ref{tab:attention_visualization}.
We noticed that our model learned to focus on technical terms, such as \textit{lspci}, \textit{shutdown}, and \textit{traceroute}.
We also observed that the model is able to capture contextual importance, i.e.~it is able
{to focus on context relevant words.}
For example, given the context in Table~\ref{tab:context} and the correct response in the  
first row {of} Table~\ref{tab:attention_visualization},
{one can see the attention on the word {\emph{shutdown}}, where it gets a lower weight when used as a common word in the first occurance than 
as a UNIX command in the second.~\footnote{N.B. The conversations are taken directly from the corpus and can be grammatically inconsistent.}} 


\begin{table}[t]
    \centering
    \begin{tabularx}{\linewidth}{>{\raggedright\arraybackslash}X}
         \toprule
         \textbf{Context utterances} \\
         \cmidrule{1-1}
         \textbf{Utterance 1}: Ubuntu $<$version$>$ \\
         \textbf{Utterance 2}: hi all sony vaio fx120 will not turn off when shutting down, any ideas? btw acpi =o ff in boot parameters anything else i should be trying?  \\
         \textbf{Utterance 3}: how are you shutting down i.e. terminal or gui? \\
         \bottomrule
    \end{tabularx}
    \caption{Sample context utterances from UDC's test set whose correct response is the first utterance in Table~\ref{tab:attention_visualization}.}
    \label{tab:context}
\end{table}

\begin{table}[t]
    \centering
    \begin{tabularx}{\linewidth}{>{\raggedright\arraybackslash}X}
        \toprule
         \textbf{Examples of model error:} \\
         \cmidrule{1-1}
         \textbf{Correct}: ok will do :), nope.  \\
         \textbf{Predicted}: \_\_url\_\_ if you go down to the bottom of that tutorial i also have a post there that is a bit more detailed about my problem poster name is trent \\  
         \cmidrule{1-1}
         \textbf{Correct}: hmm! ok  \\
         \textbf{Predicted}: as did i w/ fbsd ... just check out the livecd for a bit \\  
         \cmidrule{1-1}
         \textbf{Correct}: okay thank you a thread i hope :)  \\
         \textbf{Predicted}: hmm ok because im not sure about iwconfig and wpa but we can give it a try do gksudo gedit \_\_path\_\_ then add a record like this \_\_url\_\_ \\  
         \cmidrule{1-1}
         \textbf{Correct}:  right .. it is, it exists i verified  \\
         \textbf{Predicted}: i want to connect to your computer remotely if you allow me to so i can fix the problem for you just follow the following procedure. \\  
         \cmidrule{1-1}
         \textbf{Correct}: roger .. lemme check, got it ... thanks dude :)  \\
         \textbf{Predicted}: just click the partition and then click the blue text next to mount point or you can simply navigate to that path \\   
        \bottomrule
    \end{tabularx}
    \caption{{Examples on the error our model made. We observed that our model's predictions are biased towards non-generic responses.}}
    \label{tab:error}
\end{table}
    
\subsection{Error analysis}


We qualitatively analyzed the errors our method made. We observed that our model's predictions are biased toward high information utterances. That is, {we observed for some examples that} the correct response {is} generic (i.e.~{has} low information), our model chooses {a} non-generic response, as shown in Table~\ref{tab:error}. Furthermore, we computed the average utterance information content ({the} entropy) {for} both the correct and predicted responses, based on~\citet{xu2018information}, {where we 
obtained} 9.25 bits and 9.34 bits, 
respectively. This quantitatively indicates that our model is slightly biased toward high information responses.

\section{Conclusion and future work}

We presented a novel model 
{which extends} the dual encoder architecture for multi-turn response selection by incorporating external domain knowledge and attention augmented encoding. Our experimental results {demonstrate} that our model outperformed other state-of-the-art methods for response selection in a multi-turn dialogue {setting}, 
{and}
that {the} attention mechanism and {incorporating} additional domain knowledge are indeed effective approaches for improving the {response selection performance of the} dual encoder architecture. 
Further improvement {might} be made by also considering domain knowledge in the context and by {improving the} handling {of} OOV {words}, e.g.~{by} widening our domain specific word vocabulary and handling generic OOV words such as typos.

\bibliography{conll2018}
\bibliographystyle{acl_natbib_nourl}

\appendix

\end{document}